\documentclass[final]{midl} 

\usepackage{mwe} 

\newcommand{\erik}[1]{\textcolor{black}{#1}}

\title[WSSS Ensembles for Medical Imaging]{Exploring Weakly Supervised Semantic Segmentation Ensembles for Medical Imaging Systems}

\midlauthor{\Name{Erik Ostrowski} \Email{erik.ostrowski@tuwien.ac.at}\\
\Name{Bharath Srinivas Prabakaran} \Email{bharath.prabakaran@tuwien.ac.at}\\
\addr Institute of Computer Engineering, Technische Universit{\"a}t Wien (TU Wien), Austria\\
\Name{Muhammad Shafique} \Email{muhammad.shafique@nyu.edu}\\
\addr Division of Engineering, New York University Abu Dhabi, United Arab Emirates\\
}

\begin{document}

\maketitle

\begin{abstract}
Reliable classification and detection of certain medical conditions, in images, with state-of-the-art semantic segmentation networks, require vast amounts of pixel-wise annotation.
However, the public availability of such datasets is minimal.
Therefore, semantic segmentation with image-level labels presents a promising alternative to this problem.
Nevertheless, very few works have focused on evaluating this technique and its applicability to the medical sector.
Due to their complexity and the small number of training examples in medical datasets, classifier-based weakly supervised networks like class activation maps (CAMs) struggle to extract useful information from them.
However, most state-of-the-art approaches rely on them to achieve their improvements.
Therefore, we propose a framework that can still utilize the low-quality CAM predictions of complicated datasets to improve the accuracy of our results.
Our framework achieves that by first utilizing lower threshold CAMs to cover the target object with high certainty;
second, by combining multiple low-threshold CAMs that even out their errors while highlighting the target object.
We performed exhaustive experiments on the popular multi-modal BRATS and prostate DECATHLON segmentation challenge datasets.
Using the proposed framework, we have demonstrated an improved dice score of up to 8\% on BRATS and 6\% on DECATHLON datasets compared to the previous state-of-the-art.
\end{abstract}

\begin{keywords}
Semantic Segmentation, CAMs, Deep Learning, Image-level Labels.
\end{keywords}


\section{Introduction}

Semantic segmentation describes classifying objects in a given image and returning their pixel-wise locations.
Semantic segmentation offers, in contrast to simple classification, a particular region of detection within the image, therefore offering a much higher degree of explainability of the network's decision.
Therefore, their use in medicine would help a doctor or clinician conduct further investigations based on the assistant's decision.

State-of-the-art models for semantic segmentation are achieving new top performances at incredible speed, enabling this approach in the field of computer-aided diagnostics (CAD) and other related to medicine.
For example, there are a variety of different applications, like colon crypt segmentation \cite{COHEN2015150}, brain tumor detection \cite{tek2002segmentation,rehman2020deep}, detection of gastric cancer \cite{an2020deep, song2020clinically}, discovering and tracking medical devices in surgery\cite{wei1997automatic, zhao2017tracking}, etc..
However, one significant factor limiting the quality of semantic segmentation results in medicine is that state-of-the-art methods are fully supervised.
Fully supervised means that to achieve high-quality prediction results, the models require vast training data with pixel-wise annotated ground truth masks.
Such datasets almost do not exist in the medical field because generating those masks in a normal context is tedious and time-consuming \cite{cordts2016cityscapes}.
However, in the medical sector, this task gets exponentially more difficult.
Hence, other solutions are needed if we want to leverage the power of semantic segmentation networks in the medical sector on a broader level.
One such approach is to create semantic segmentation with only image-level annotations.
\erik{We focus only on semantic segmentation using image-level annotations in this work, because image-level annotations are the fastest and easiest to acquire, compared to any other weaker form of supervision}

Several past works in this field have already shown that this approach can lead to promising results \cite{du2021weakly, jiang2022l2g, chang2020weakly}.
However, their applications in the field of medicine are still very niche.
Patel et al. \cite{PATEL2022102374} have tried to apply those methods specifically in a medical context.
Their WSS-CMER approach uses a class activation map-based approach (CAM) that extends the ideas proposed by \cite{wang2020self}.
CAMs are the most popular method for predicting semantic segmentation using image-level labels.
Furthermore, Guo et al.\cite{guo2020weakly} used image-level labels to localize organs in 3D images.
\cite{chen2022c} uses an adapted CAM approach for image-level segmentation on image in medical context.
Nevertheless, CAMs have been shown to be reliable for the localization of target objects, but their masks lack detail.
For example, with a brain tumor, they will highlight the center of the tumor but become more uncertain towards the edges, either including too many background pixels or not covering the whole tumor.
Therefore research concentrates on using CAMs as a base prediction and then aims to construct methods to improve their predictions. 

In this light, the state-of-the-art approaches mostly tackle the classification loss of the CAM by adding regularizations that aim to guide the network to predict finer and more complete masks.
For example, additional regulations for the loss function were used in \cite{PATEL2022102374} and \cite{wang2020self}, like matching the predictions of affine transformations to the original predictions.
Alternatively, many approaches refine the CAM prediction after the fact, using methods like pixel-similarity, for example, as in \cite{ahn2018learning}.
Nevertheless, those approaches do not work as well in the medical sector.
We observed that we were unable to generate reliable classifiers on our target datasets due to size and complexity.
Therefore, using the adapted loss functions, like in \cite{wang2020self}, on top of the classification loss does not improve the result.
Moreover, we also observed that refining the CAM masks after the fact was not easy since their quality was too low.
Based on our initial experiments, we have made the following observations: First, CAM predictions with a low enough threshold cover the target object with very high confidence.
Second, when examining the output of different CAMs, be it different base models or altogether different methods, we noticed that they vary, especially when applying low thresholds.
Therefore, we propose a framework that combines the predictions of multiple Grad-CAMs \cite{selvaraju2017grad} built using different base models by determining the most optimal ensemble of thresholds in the training set.
We conducted an exhaustive analysis of our framework on the popular BRATS 2022 \cite{menze2014multimodal, bakas2017advancing, bakas2018identifying} and DECATHLON \cite{antonelli2022medical} datasets and compared it with the relevant state-of-the-art approaches in this context.
Our main contributions can be summarized as follows:
\begin{enumerate}
    \item A novel semantic segmentation framework using just medical imaging labels by evaluating ensemble network configurations. 
    \item Evaluations on the BRATS and DECATHLON datasets have achieved performance improvements of 6\% and 8\%, respectively, over the current state-of-the-art. 
    \item Our framework is completely open-source and accessible online\footnotemark.
    
\end{enumerate}

\footnotetext{\url{https://github.com/ErikOstrowski/Automated_Ensemble}} 

\section{Our Automated Ensemble Search}

\begin{figure}[htbp]
\floatconts
  {fig:framework}
  {\caption{Overview of our Framework. 
(1) Stages in the classifier training, which will be used in the Grad-CAM;
(2) The Grad-CAM uses trained models and input images to generate CAMs; 
(3) The Ensemble will test out multiple methods of how to combine the CAMS; }}
  {\includegraphics[width=0.9\columnwidth]{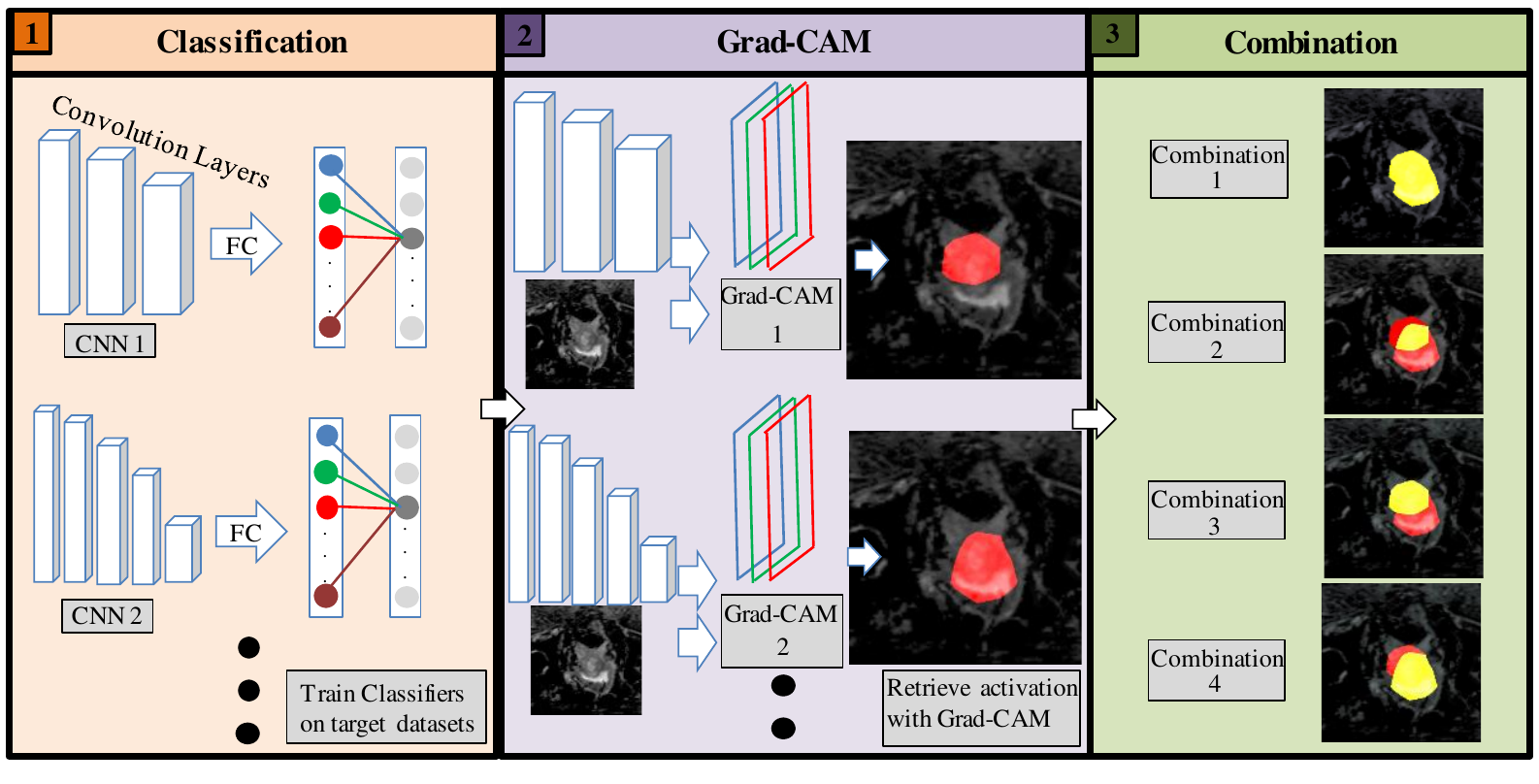}}
  \label{framework}
\end{figure}

Fig.~\ref{fig:framework} presents an overview of our framework.
First, we start by training a classifier model on the target dataset.
In our case, the ResNet-34 and ResNet-50 \cite{he2016deep} models proved to be the most successful, but the framework accommodates the use of any other classifier instance. 
Second, we use Grad-CAM to create the first masks
 for the different classifiers.
In our case, the standard Grad-CAM worked the best.
Next, we test ensemble methods to combine two or more prediction sets.
For our final version, we choose the ensemble version that offers the best possible results, 
followed by the calibration step to determine the best ensemble of thresholds for the highest detection score.

\subsection{Training and Exploration of Classifier Model Instances}

Our framework aims to create an ensemble of different CAM methods because their ensemble evens out their shortcomings and therefore results in more accurate predictions than their singular pieces.
Following the observation that the best ensembles are generated from high-quality CAMs, and high-quality CAMs are generated from high-quality classifiers, we have to investigate classifiers for our target datasets.
We limited ourselves to multiple ResNets classifiers in our experiments.
Note that in our framework, any classifier can be used. 
\erik{Additional information about the ResNets can be found in Appendix~\ref{sec:res}.}

Instead of striving to test more complex networks, we took some inspiration from other methods like the self-supervised Swav \cite{caron2020unsupervised}.
Swav tries to learn to differentiate pictures without guidance instead of using annotations.
For this purpose Swav uses a contrastive loss function that compares pairs of images.
The goal of the loss function is to push away images that are different in the feature space while pulling together those from transformations, or views, of the same image in the feature space.
In particular, our interest in the Swav approach relies on two reasons: first, using a pre-trained unsupervised model for the medical datasets may improve the quality of the Grad-CAM results.
This may be the case since unsupervised models are guided more toward distinguishing between shapes rather than just guessing the correct class.
Second, many state-of-the-art approaches add additional regularization to the classification loss. 
Those regularizations, e.g., the affine transformation in SEAM, are very much in the same spirit as Swav's contrastive learning loss.
Therefore, we assume that the activations of an unsupervised trained model may recognize the objects more completely.
We have evaluated multiple trained Swav models but observed that their contrastive loss approach did not result in higher-quality CAMs compared to the conventionally trained classifiers.

\subsection{Evaluating Grad-CAMs for the trained models}
After creating our candidate classifiers, we can focus on generating CAM predictions.
For that purpose, we will apply Grad-CAMs.
Gradient-weighted Class Activation Mapping (Grad-CAM) takes a network and image as input and returns a rough mask.

However, it was shown that Grad-CAM fails to properly localize objects in an image if it contains multiple occurrences of the same class.
Moreover, it was also found that due to not weighing the average of the partial derivatives, the localization often does not correspond to the entire object but only parts of it.
Hence, Grad-CAM++ \cite{chattopadhay2018grad} was introduced, which solves those problems by using a more sophisticated formula for the importance score.
As a further optimization, SmoothGrad-CAM++ \cite{omeiza2019smooth} was introduced.
\erik{Additional information about GradCAMs can be found in Appendix~\ref{sec:grad}.}
However, these improvements aim to increase the sharpness of the boundaries of the detected object and alleviate the issues of the original method with multiple objects of the same class within the same image.
But, many of those problems do not occur in the observed medical datasets.
All images in the BraTS and DECATHLON datasets never contain multiple instances of the target object. 
Furthermore, those target objects often have a round-ish shape, whose borders are ambiguous even for experts.
Nevertheless, we also tested SmoothGrad-CAM++ as it is a more recent iteration of this approach.
For CAM generation, we will run our trained candidate models and the images through the candidate Grad-CAM, creating masks for all images.

\subsection{Ensemble methods}
\begin{figure}[htbp]
\floatconts
  {fig:ensemble}
  {\caption{Visualization of the ensemble methods: (a) `or', (b) `and', (c) `$<$', (d) `$>$'}}
  {\includegraphics[width=0.9\columnwidth]{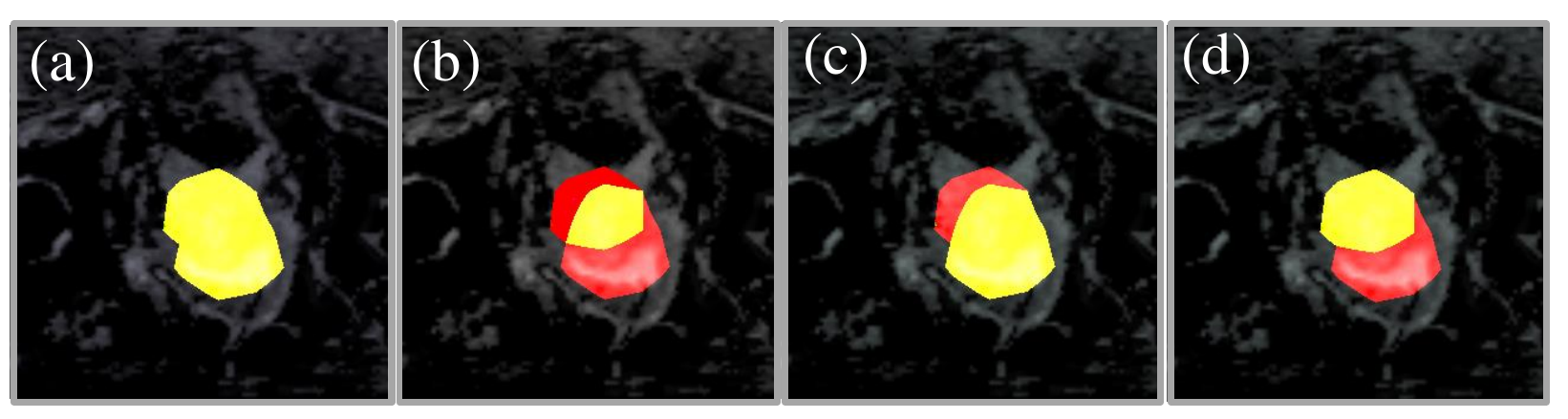}}
  \label{ensemble}
\end{figure}

Once we have the collected masks of our candidate models and Grad-CAMs, we aim to combine them for higher-quality results.

We considered four simple and intuitive approaches for the ensemble methods, Fig. \ref{ensemble} presents a visualization of those methods.
First, we have the `or' ensemble, which sums up the predictions of candidates.
This approach works best when both masks have high true positive rates, generating the biggest possible activation area between the combined masks.
Second is the `and' approach, which multiplies the predictions of candidates.
The `and' approach minimizes the possible detection area of both masks in contrast to the `or' approach.
This approach works best when both models have a high true negative rate.
Next, the `min' and `max' approaches only take the masks with the least or most positive classified pixels, respectively.
In this way, we address the problem of models tending to predict the size of the target object as too big or too small while the complete prediction of one candidate is still better than the `and' or `or' of all candidates.

Since the Grad-CAM methods return the predictions in logits ranging from $0$ to $1$, we can determine the threshold of the value at which we consider the pixel a positive or negative classification.
This hyperparameter gives us a massive leeway about any given prediction's false positive and false negative rate.
Using very high thresholds drastically reduces the area classified as positive, leading to a high false negative rate.
And vice-versa, using very low thresholds drastically increases the area classified as positive, leading to a high false positive rate.
The most optimal threshold value varies from candidate model to candidate model.
Therefore, we decided to test the chosen ensemble approach with all combinations of thresholds from $0$ to $1$ in steps of $0.1$.
We conduct these tests on the training set to determine the thresholds we would use on the validation set. 
Our experiments have shown that the combination of thresholds most optimal for the training set is also one of the most optimal combinations for the validation set.
A more comprehensive ablation study of all the threshold combinations used can be found in Appendix~\ref{sec:Thresh}.

\section{Results and Discussion}
\label{sec:mainRes}

We compare the results of our framework to relevant prior literature, namely WSS-CMER and SEAM, which was a baseline used in the WSS-CMER work.
To the best of our knowledge, WSS-CMER is the only method in this domain specifically targeted for medical images.
Moreover, we include the results of SEAM because it represents a current approach to weakly supervised segmentation that is not optimized for data from the medical sector.
The goal of our framework was to extend current applications of semantic segmentation with image-level labels to the medical sector. 
Primarily because we notice a massive demand for dealing with data scarcity in the medical sector, especially with respect to the availability of pixel-wise annotated medical images. 


First, we discuss the experimental setup used for the experiments discussed in this section. 
We ran the experiments on a CentOS 7.9 Operating System executing on an Intel Core i7-8700 CPU with 16GB RAM and 2 Nvidia GeForce GTX 1080 Ti GPUs.
We executed our scripts with the following software versions: CUDA 11.5, Pytorch 1.13.0, and torchvision 0.14.0.
We tested our framework on the BraTS 2020 and DECATHLON datasets for evaluation and comparison.
Additional information about the evaluation metrics used can be found in Appendix~\ref{sec:Eval}.

Moreover, we trained and evaluated the datasets in a three-fold cross-validation scheme.
For BraTS, we divided the randomly shuffled dataset into three equally sized batches and trained and tested three models, each trained on two of the three samples and evaluated on the remaining sample.
For DECATHLON, we downloaded the three-fold cross-validation splits of WSSS-CMER.
We reported the average results over the three runs.
Note that we could not reproduce the results of WSS-CMER and SEAM.
Therefore we used the numbers published in \cite{PATEL2022102374}, \erik{when stating the results from WSS-CMER and SEAM.
We ended up using mostly the ensemble of ResNet-34 and ResNet-50 as those two base models achieved the be results on their own.
Adding more components or different models to the ensemble resulted in lower prediction quality.}
A visualization of the results on example images can be found in Appendix~\ref{sec:Res} for both datasets.

\textbf{BraTS:}
The BraTS 2020 dataset is a multi-modal brain tumor segmentation in Magnetresonance images.
Three hundred sixty-nine multi-modal scans with their corresponding expert segmentation masks are available for the tumor segmentation task.
For comparability with WSS-CMER and SEAM, we will exclude the additional 24 scans of BraTS 2020. 
These datasets use the same naming convention and direct filename mapping between them.
The scans are composed of four modalities, which include T1, T1c, T2, and T2 Fluid Attenuated Inversion Recovery (FLAIR).
We use T1c, T2, and FLAIR for classifier training, and for mask generation with Grad-CAM, we only use FLAIR.
We achieved the best results in our experiments just using FLAIR .
We separated each scan into its set of slides, resulting in 47,232 slides, cropped them to the size $128\text{x}128$, and applied a threshold of $0.8$ to the normalized frame, which generated the best results.

For this work, we followed the procedure of WSS-CMER, which only considered a binary segmentation class, i.e., healthy vs. non-healthy targets.
Therefore, we merge the different tumor classes into one single `positive' class.
Our framework is not limited to binary scenarios.
However, since the used datasets are quite complex and the work on this domain in the medical sector is still in the early stages, we are constrained to this simplified scenario.
Nevertheless, Grad-CAM predictions have been shown to be class-specific in natural images, as widely observed in the computer vision literature.
But, in natural images, we do not find the case that certain classes are only present in combination with other classes, which enables the classifier to discriminate between those classes.
In both of the used datasets, most classes only appear in combination with the other classes, so our classifier cannot distinguish between them.
Since we observed that our classifiers are already struggling to distinguish between images with or without the presence of a tumor, we saw no point in adding different tumor classes additionally.
For instance, \cite{wu2019weakly} reported that adding those classes results in worse predictions.
In table \ref{tab2}, we compare the results of the BraTS dataset.
\erik{We added the “Empty” mask, where we just predicted a background mask for every image.
Note that \cite{PATEL2022102374} do not achieve an improvement over “Empty” masks, which stresses the necessity to improve predictions in this use case.
We also added the performance of a fully supervised semantic segmentation (FSSS) network for both datasets as an upper bound.
When we tried to reproduce \cite{PATEL2022102374}'s results, we experienced that the classifiers had an accuracy of around $50\%$ for both datasets, which is in line with all other classifiers that we tested.
This shows, that adding a more sophisticated loss to the training process, like in WSS-CMER or SEAM, does not lead to better CAMs, if the baseline classifier is not able to work with the respective dataset.}
Our baseline Grad-CAM with ResNet-34 is around $7\%$ better than WSS-CMER.
Moreover, Grad-CAM ResNet-50 is  $1\%$ better than its ResNet-34 counterpart.
Using the bigger ResNet-101 did achieve worse results, as anticipated, due to its larger number of parameters. 
We did not observe a difference in CAM quality when using SmoothGrad-CAM++ for both datasets.
Finally, by combining the two ResNets via the `and' scheme, we generate our best result, which is $1.5\%$ better than the previous high score.
\erik{Note that SEAM and WSS-CMER showed a variation of over $11\%$ between samples, while our Ensemble varied less than $2\%$.
Additional information about the variations between the sample evaluations can be found in Appendix~\ref{sec:var} in table\ref{tabf2}.}

\textbf{DECATHLON:}
The DECATHLON segmentation challenge is a multi-modal prostate dataset, with the task to localize and detect the prostate peripheral zone and the transition zone.
The dataset consists of 32 volumetric scans containing MRI-T2 and apparent diffusion coefficient (ADC) maps with their corresponding segmentation masks.
We only used the ADC maps for classifier training and CAM generation, as they generated the best results.
We followed a similar scheme as with BraTS, where the prostate peripheral zone and the transition zone are combined into a single `positive' class.
In table \ref{tab1}, we compare the results of the DECATHLON datasets.
Our baseline Grad-CAM with ResNet-34 is around $7\%$ better than WSS-CMER.
However, Grad-CAM ResNet-50 is $2\%$ worse than its ResNet-34 counterpart.
The smaller ResNet-18 achieved worse results as it could not learn the relevant information.
Finally, by combining the two ResNets via the `and' scheme, we generate our best result, which is $1.4\%$ better than the previous high score.
\erik{Note that SEAM and WSS-CMER showed a variation of over $13\%$ between samples, while our Ensemble varied less than $3.5\%$.
Additional information about the variations between the sample evaluations can be found in Appendix~\ref{sec:var} in table\ref{tabf1}.}

\begin{table}[t]
\centering 
{
\begin{tabular}{lcc}
\hline
\bfseries Method        & \bfseries Best AVG DSC & \bfseries Best AVG mIoU \\ \hline

SEAM                  & 56.1 & 39.0\\
WSS-CMER              & 59.7 & 42.6\\
Empty                 & 64.6 & 47.7 \\
ResNet-34             & 67.3 & 50.7 \\
ResNet-50             & 68.5 & 52.1 \\ \hline
Ensemble ResNet-34x50 `or'      & 67.8 & 51.3\\
Ensemble ResNet-34x50 `$<$'       & 68.3 & 51.9\\
Ensemble ResNet-34x50 `$>$'       & 68.6 & 52.2\\ 
Ensemble ResNet-34x50 `and'     & \textbf{70.3} & \textbf{54.2}\\\hline
FSSS                  &81.8 &69.2\\ \hline
\end{tabular}}
\caption{Comparison of EnsembleCAM with state-of-the-art techniques on the BraTS dataset.\label{tab2}}
\end{table}

\begin{table}[t]
\centering 
{
\begin{tabular}{lcc}
\hline
\bfseries Method         & \bfseries Best AVG DSC & \bfseries Best AVG mIoU \\ \hline

VGG19                 & 63.6 & 46.6 \\
Empty                 & 65.5 & 48.7 \\
ResNet101                 & 65.9 & 49.1 \\
SEAM                  & 65.9 & 49.1 \\
ResNet-18                 & 67.0 & 50.3 \\
Swav                 & 70.2 & 54.1 \\
WSS-CMER              & 71.3 & 55.4\\
ResNet-50             & 76.1 & 61.4 \\
ResNet-34             & 78.0 & 63.8 \\ \hline
Ensemble ResNet-34x50 `or'      & 77.3 & 63.0\\
Ensemble ResNet-34x50  `$<$'      & 77.7 & 63.5\\
Ensemble ResNet-34x50 `$>$'       & 78.4 & 64.4\\
Ensemble ResNet-34xSwav `and'     & \textbf{78.7} & \textbf{64.9} \\
Ensemble ResNet-34x50 `and'     & \textbf{79.3} & \textbf{65.7} \\ \hline
FSSS                  &86.8 &76.7\\ \hline

\end{tabular}}
\caption{Comparison of EnsembleCAM with state-of-the-art techniques on the DECATHLON dataset.\label{tab1}}
\end{table}

\section{Conclusion}
In this paper, we have proposed our novel framework, which illustrates the approach for finding an ensemble of CAM methods for semantic segmentations with image-level labels for data from the medical sector to address the lack of research specified for this application.
Our framework proposes a scheme to generate useful prediction masks despite the lack of quality prediction of base models due to the complexity and size of the used datasets.
Therefore, this framework can also be applied in different contexts where the standard approaches do not generate high-quality results.
We showed that the predictions generated by our framework achieve state-of-the-art performance on the BraTS 2020 and DECATHLON datasets, which proves its effectiveness compared to other approaches.
Our framework is open-source and accessible at \url{https://github.com/ErikOstrowski/Automated_Ensemble}.

\midlacknowledgments{This work is part of the Moore4Medical project funded by the ECSEL Joint Undertaking under grant number H2020-ECSEL-2019-IA-876190.}

\bibliography{midl-samplebibliography}

\appendix

\erik{
\section{ResNets}
\label{sec:res}
When we aim to use deeper architectures than VGG, we must resort to networks that employ techniques like ResNet's skip connections to avoid the vanishing gradients problem.
Additionally, ResNet introduced batch normalization as a regularization strategy.
In contrast to VGG's \cite{simonyan2014very} 19 layers, ResNet can utilize up to 200.
Hence, ResNet can retrieve more complex information and is better at handling larger datasets.
Nevertheless, in our experiments, we encountered worse results when employing bigger ResNets, and therefore, we refrained from employing deeper and more complex networks.
}

\erik{
\section{GradCAMs}
\label{sec:grad}
GradCAM runs an input image through a model and takes the $K$ outputs  $ A^k \in \mathbb{R}^{u \times v} $ of the final convolutional layer.
Next, we calculate the gradient score for each class $c$, of the logits $y^c$ for the feature map activations $A^k$, i.e., $\frac{\partial y^c}{\partial A^k}$.
Then the gradients are global average pooled across each feature map to give us the importance score $ \alpha^c_k$:
$$ \alpha^c_k = \frac{1}{uv} \sum_i \sum_j \frac{\partial y^c}{\partial A^k_{i,j}}$$
where $k$ denotes the index of the $Z$ activation maps, and $i,j$ are the feature map coordinates.
The importance score weights the relevance of each feature map $k$ for each class $c$.
Then, we multiply each activation map $A^k$ by its importance score $\alpha^c_k$ for each class and take their sum to gain the prediction mask $M^c$:
$$M^c = ReLU(\sum_k \alpha^c_k A^k)$$
We apply ReLU to the summation to only consider the pixels that positively influence the score of the class of interest. 
The mask  $M^c$ highlights areas of the image that were activated for the respective class.
Finally, we resize and rescale $M^c$ so that the mask fits the original image dimensions.
GradCAM++'s formula for the importance score $\alpha_{i,j}^{c_k}$:
$$ \alpha_{i,j}^{c_k} = \frac{\frac{\partial^2 y^c}{(\partial A^k_{i,j})^2}}{2 \frac{\partial^2 y^c}{(\partial A^k_{i,j})^2}+ \sum_a \sum_b A_{ab}^k ( \frac{\partial^3 y^c}{(\partial A^k_{i,j})^3})}$$
SmoothGrad-CAM++ combined Grad-CAM++ with SMOOTHGRAD \cite{smilkov2017smoothgrad}, which improves the original method by sharpening the Grad-CAM output by taking random samples in a neighborhood of an input $x$ and averaging the resulting outputs.
}

\section{Evaluation metrics}
\label{sec:Eval}

To assess the performance of the proposed approach, we employ the common Dice Similarity coefficient (DSC) and mean Intersection over Union (mIoU):
$$DSC(GT,Pred) = \frac{2 | GT \cdot Pred|} {|GT|+|Pred|}$$
Where $GT$ and $Pred$ are binary matrices (with values of 1 for elements inside a group and 0 otherwise), $GT$ signifies the ground truth, and $Pred$ signifies the classification result.\\
The mean Intersection-over-Union (mIoU):
$$mIoU = \frac{1}{N} \sum_{i=1}^N \frac{ p_{i,i}}{\sum_{j=1}^N p_{i,j} + \sum_{j=1}^N p_{j,i} - p_{i,i}  }$$

\noindent where $N$ is the total number of classes, $p_{i,i}$ the number of pixels classified as class $i$ when labelled as class $i$. $p_{i,j}$ and $p_{j,i}$ are the number of pixels classified as class $i$ that were labelled as class $j$ and vice-versa, respectively.

\section{Analyzing results of the framework}
\label{sec:Res}


Fig.~\ref{fig:example} presents an overview of our experimental results.
The first two rows show examples from the DECATHLON dataset, and the second two rows show examples from the BraTS dataset.
We notice that the (f) with the `and' combination achieves the biggest overlap with the ground-truth row (b) in terms of visual and metric-based evaluations, as shown in Section~\ref{sec:mainRes}.
Whereas in contrast, the `or' ensemble used in the (e) row creates the results with the least amount of overlap when considering the false positives.
This shows our observation that motivated the framework. 
Both ResNet-34 and ResNet-50 cover a lot of ground-truth but also too many background pixels.
However, we notice that the activated background pixels differ between the two models, resulting in the best predictions when using the `and' ensemble and the worst when using the `or' ensemble.

\begin{figure}[t]
\floatconts
  {fig:example}
  {\caption{Example results of our Framework on BraTS and DECATHLON:
  (a) Source image, (b) ground-truth, (c) ResNet34 Grad-CAM, (d) ResNet50 Grad-CAM, (e) `or' ensemble, (f) `and' ensemble}}
  {\includegraphics[width=\columnwidth]{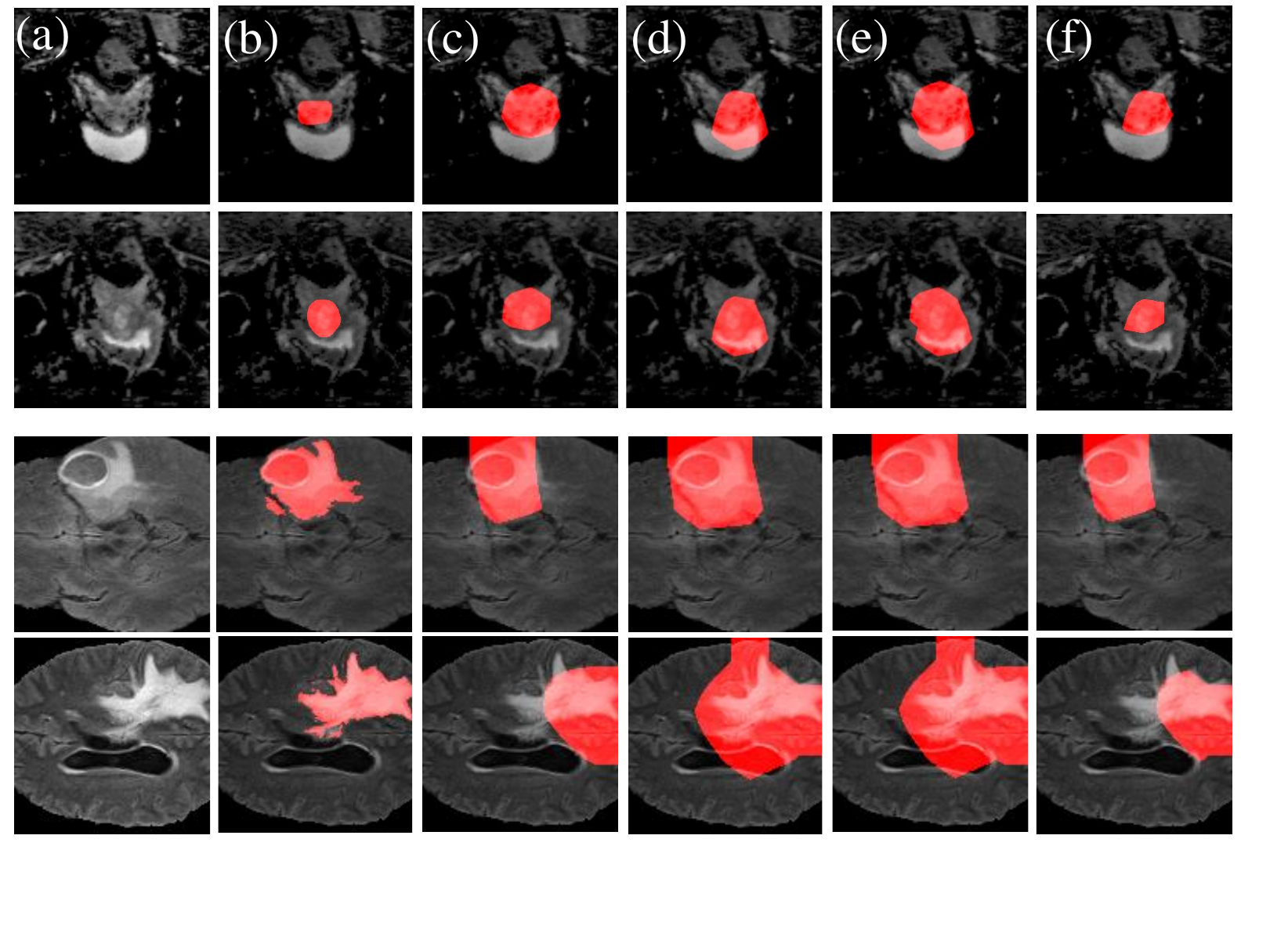}}
  \label{example}
\end{figure}

\section{Threshold testing}
\label{sec:Thresh}

Fig.~\ref{fig:heat} presents an overview of the mIoU achieved when different thresholds are used for the two models under consideration. 
These values are obtained by using the training dataset results averaged over all three cross-validation iterations.
The left graphic shows the results of BraTS, and the right shows the DECATHLON results.
We notice that the left graphic is lighter in color than the right.
The reason for that is that overall mIoU results on BraTS are lower than the results on DECATHLON.
We notice that both datasets reach their best results at thresholds $70\%$ for both models. 
Both graphics show results achieved with the `and' ensemble; the optimal thresholds appear to vary based on the ensemble method under consideration. 

\begin{figure}[t]
\floatconts
  {fig:heat}
  {\caption{All threshold combinations of ResNet-34 and ResNet-50 `and' combination on the train set.
  Y-axis describes the threshold used for ResNet-34, X-axis is the threshold used for ResNet-50. BraTS on the left, DECATHLON on the right.}}
  {\includegraphics[width=\columnwidth]{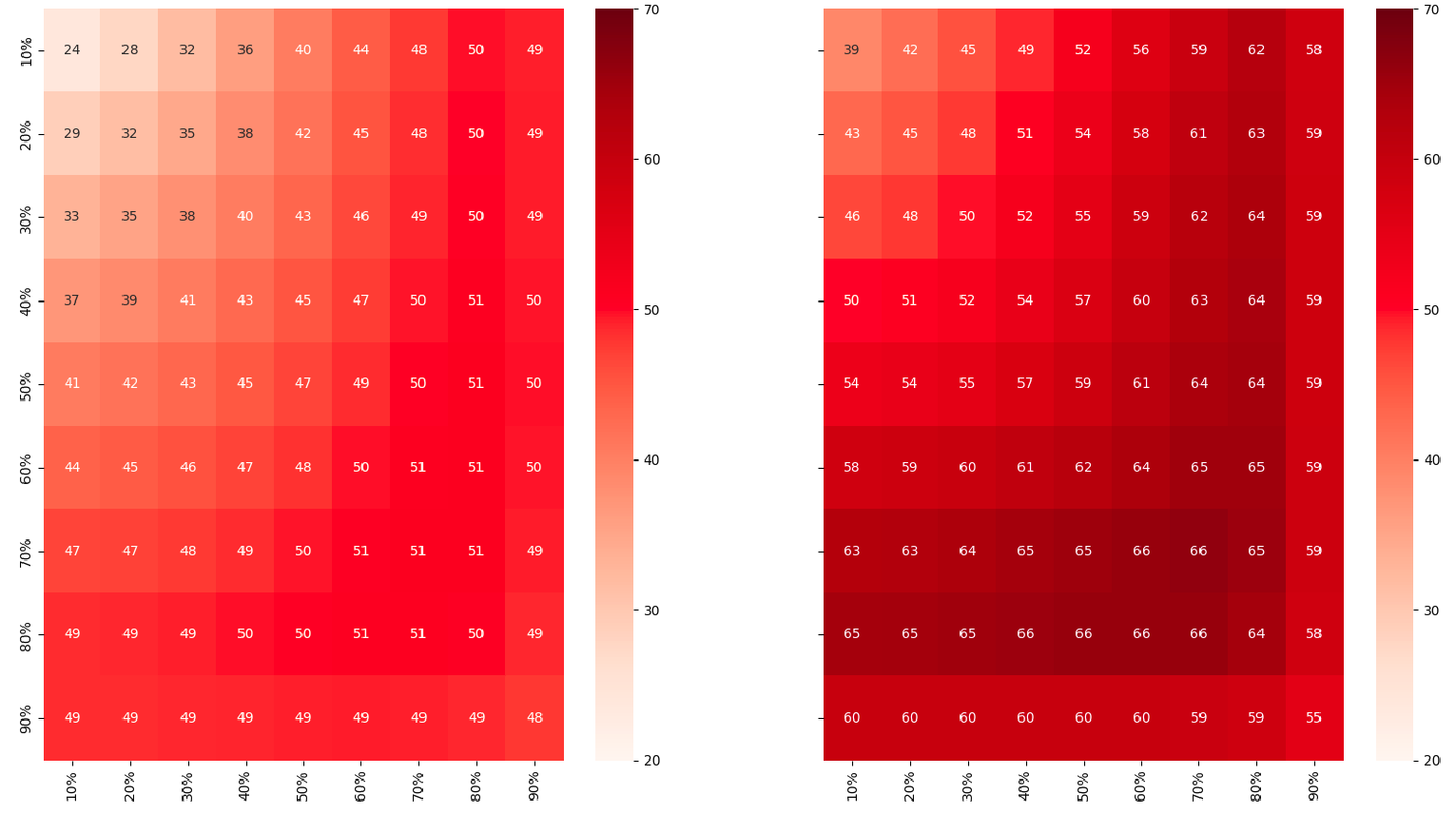}}
  \label{heat}
\end{figure}

\section{Analyzing Sample Variations}
\label{sec:var}

\begin{table}[t]
\centering 
{
\begin{tabular}{lcc}
\hline
\bfseries Method        & \bfseries Best AVG DSC & \bfseries Best AVG mIoU \\ \hline

SEAM                  & 56.1 $\pm$ 11.9 & 39.0 $\pm$ 6.3\\
WSS-CMER              & 59.7 $\pm$ 11.8& 42.6 $\pm$ 6.3\\
Ensemble ResNet-34x50 `and'     & \textbf{70.3} $\pm$ 1.5 & \textbf{54.2} $\pm$ 0.8\\\hline
\end{tabular}}
\caption{Comparison of EnsembleCAM with state-of-the-art techniques on the BraTS dataset, including the variations observed between the samples.\label{tabf2}}
\end{table}

\begin{table}[t]
\centering 
{
\begin{tabular}{lcc}
\hline
\bfseries Method         & \bfseries Best AVG DSC & \bfseries Best AVG mIoU \\ \hline

SEAM                  & 65.9 $\pm$ 13.35 & 49.1 $\pm$ 7.2\\
WSS-CMER              & 71.3 $\pm$ 14.6& 55.4 $\pm$ 7.9\\
Ensemble ResNet-34x50 `and'     & \textbf{79.3} $\pm$ 3.1 & \textbf{65.7} $\pm$ 1.6  \\ \hline

\end{tabular}}
\caption{Comparison of EnsembleCAM with state-of-the-art techniques on the DECATHLON dataset, including the variations observed between the samples.\label{tabf1}}
\end{table}

Table.~\ref{tabf2} and table~\ref{tabf1} give an overview of the DSC and mIoU achieved with state-of-the-art techniques on both datasets, including the variations between the three different validation samples.
Note that we cannot give a more comprehensive analysis regarding what exact score was achieved on each validation set, because the WSS-CMER did only provide the average variations.
Nevertheless, we notice that for both datasets, the variation in the evaluation results is much lower for EnsembleCAM compared to both SEAM and WSS-CMER.
Showing that EnsembleCAM is much more consistent when being confronted with new data.

\end{document}